\documentclass[journal]{IEEEtran}

\usepackage{color}

\usepackage{soul}
\usepackage{authblk}
\usepackage{gensymb}
\usepackage[font=footnotesize]{caption}
\renewcommand{\baselinestretch}{.95}
\usepackage{graphicx}
\usepackage{subfigure}

\ifCLASSINFOpdf
\else
\fi

\usepackage[pscoord]{eso-pic}

\usepackage{amsmath, amssymb, amsthm}
\usepackage{array}
\usepackage{color}
\usepackage{subfigure}
\usepackage{tikz}
\usepackage{textcomp}
\usepackage{hyperref}
\usepackage{lipsum}

\begin{document}

\title{Motion Corrected Multishot MRI Reconstruction Using Generative Networks with Sensitivity Encoding}
\author[1]{Muhammad Usman}
\author[1]{Muhammad Umar Farooq}
\author[2]{Siddique Latif}
\author[1]{Muhammad Asim}
\author[1]{Junaid Qadir}

\affil[1]{Information Technology University (ITU)-Punjab, Pakistan}
\affil[2]{University of Southern Queensland, Australia}

\maketitle

\begin{abstract}
Multishot  Magnetic  Resonance  Imaging (MRI) is a promising imaging modality that can produce a high-resolution image with relatively less data acquisition time. The downside of multishot MRI is that it is very sensitive to subject motion and even small amounts of motion during the scan can produce artifacts in the final MR image that may cause misdiagnosis. Numerous efforts have been made to address this issue; however, all of these proposals are limited in terms of how much motion they can correct and the required computational time. In this paper, we propose a novel generative networks based conjugate gradient SENSE (CG-SENSE) reconstruction framework for motion correction in multishot MRI. The proposed framework first employs CG-SENSE reconstruction to produce the motion-corrupted image and then a generative adversarial network (GAN) is used to correct the motion artifacts. The proposed method has been rigorously evaluated on synthetically corrupted data on varying degrees of motion, numbers of shots, and encoding trajectories. Our analyses (both quantitative as well as qualitative/visual analysis) establishes that the proposed method significantly robust and outperforms state-of-the-art  motion-correction techniques and also reduces severalfold of computational times 
\end{abstract}

\begin{IEEEkeywords}
Image reconstruction, magnetic resonance, motion correction, multishot acquisition, parallel imaging, Deep Learning, generative adversarial networks
\end{IEEEkeywords}

\IEEEpeerreviewmaketitle

\section{Introduction}

\IEEEPARstart{M}{agnetic} Resonance Imaging (MRI) is a safe, non-ionizing, and non-invasive imaging modality that provides high resolution and excellent contrast of soft tissues. It has emerged as a powerful and effective technique for early diagnosis of many common but potentially treatable diseases including stroke, cancer and ischemic heart disease. Despite of these advantages, the long data acquisition time of MRI causes many difficulties in its clinical as well as research applications. Numerous efforts have been proposed in the literature to expedite the data acquisition process including the use of single-shot echo planar imaging (EPI) \cite{mansfield1977multi}, parallel imaging (PI) \cite{larkman2007parallel}, and compressed sensing (CS) \cite{lustig2008compressed}).

In single-shot echo-planar imaging (EPI), all the \textit{k}-space data, necessary to reconstruct final MR image is acquired in single excitation pulse. EPI significantly accelerates the data acquisition time and minimizes the possibility of motion artifacts \cite{farzaneh1990analysis}. However, single-shot EPI suffers from low resolution and susceptibility artifacts. The stringent hardware requirements also limit the application of single-shot EPI. To overcome the limitations of single-shot EPI, segmented or multishot MRI  is used \cite{edelman1994echo}, which is an excellent compromise between echo-planar and standard spin echo imaging. 
It significantly reduces the demands on gradient performance and allows the in-plane spatial resolution to be improved to a level comparable to that of standard pulse sequences \cite{bernstein2004handbook}. However, the high-resolution volumetric imaging requires the acquisition of \textit{k}-space data with a  large  number  of  shots  at different time instances. As a result, the image may be severely degraded due to subject motion between consecutive shots. This makes the multishot sequences very sensitive to shot-to-shot variabilities caused by the motion. Therefore, the motion compensation techniques are imperatively employed to improve the quality of final MR image  in multishot MRI \cite{budde2014ultra}.

On the basis of source of motion, motion in MRI is classified into two categories. \textit{Rigid motion} is caused when some rigid part of the body such as head moves while \textit{non-rigid} motion arises from the motion of non-rigid parts of the body like arterial pulsation, cardiac motion, or by any other source in the field of view (FOV) (e.g., eyeball motion) \cite{zaitsev2015motion}. Image degradation in MR examiniation is mostly caused by rigid motion \cite{gedamu2012subject} and artifacts associated with rigid motion may cause suboptimal image quality. Subsequently, it may negatively impact radiologic interpretation \cite{brown2010prospective}, which effects the patient safety and enhances the medicolegal risks related to the interpretion of motion degraded images. Therefore, motion correction techniques are considered as an imperative part of MRI reconstruction processes. 
Previously, the problem of motion correction has been solved mostly in an iterative manner \cite{godenschweger2016motion}, which is time-consuming as well as computationally extensive.  Researchers are now increasingly interested in leveraging recent advances in machine learning (ML) and deep learning (DL) for improving the state-of-the-art in MRI motion correction. In particular, the use of generative adversarial networks (GANs) \cite{goodfellow2014generative} is interesting due to its capability of generating data without the explicit modeling of the probability density function and also due to its robustness to over-fitting. The adversarial loss brought by the discriminator formulated in GANs provides a clever way of incorporating unlabeled samples into the training and imposing higher order consistency that can be useful for motion correction in MRI. 

In this paper, we propose using a GAN-enhanced framework to correct rigid motion in multishot MRI during the brain structural scan due to its higher significance in clinical application \cite{andre2015toward}. This work is the extension of our previous preliminary work \cite{latif2018automating}, where we empirically showed the suitability of GAN for motion correction in multishot MRI. In particular, we are proposing a GAN based conjugate gradient (CG) SENSE \cite{pruessmann2001advances} reconstruction model to correct the motion in multishot MRI. The proposed techniques involve the use of CG-SENSE for the reconstruction of the motion-corrupted multishot \textit{k}-space data, which is then fed to GAN to produce an artifact-free image. The proposed technique effectively reduces the motion artifacts within significantly less amount of time which is essential for the clinical applications. Most importantly, we have validated our method on publicly available data by changing various parameters of multishot MRI such as the amount of motion, the number of shots, and the encoding trajectories. Results show that the proposed framework consistently performed better across these parameters and produces the motion-free image in significantly less reconstruction time as compared to traditional iterative techniques.

\section{Background and Related Work}
\label{Re}

MRI is highly sensitive to subject motion during the \textit{k}-space data acquisition, which can reduce image quality by inducing the motion artifacts. The artifacts by rigid motion are widely observed in multishot MR images during the clinical examination  \cite{andre2015toward}, therefore, the application of motion correction techniques is essentially performed during or after the reconstruction process to obtain an artifact-free image. Retrospective motion correction (RMC) techniques are applied to the rigid motion correction \cite{batchelor2005matrix,samsonov2010pocs}. They perform the \textit{k}-space data acquisition without considering the potential motion and object motion is estimated from acquired \textit{k}-space data \cite{zaitsev2015motion}. Many researchers proposed different RMC based method for rigid motion correction. For instance, Bydder et al. \cite{bydder2002detection} studied the inconsistencies of \textit{k}-space caused by subject motion using parallel imaging (PI) technique. The inconsistent data is discarded and replaced with consistent data generated by the parallel imaging technique to compensate the motion artifacts. This method produces an image with fewer motion artifacts albeit with a lower signal to noise ratio (SNR).

Loktyushin et al. \cite{loktyushin2013blind} proposed a joint reconstruction and motion correction technique to iteratively search for motion trajectory. Gradient-based optimization approach has been opted to efficiently explore the search space. The same authors extended their work in \cite{loktyushin2015blind} by disintegrating the image into small windows that contain local rigid motion and used their own forward model to construct an objective function that optimizes the unknown motion parameters. Similarly, Cordero et al. \cite{cordero2016sensitivity} proposed the use of a forward model to correct motion artifacts. However, this technique utilises the full reconstruction inverse to integrate the information of multi-coils for estimation and correction of motion. In another study \cite{cordero2018three}, authors extended their framework to correct three-dimensional motion (i.e., in-plane and through-plane motion). Through the plane, the motion is corrected by sampling the slices in overlapped manner.

Conventional techniques (mentioned above) estimate the motion iteratively, which makes them computationally extensive and time-consuming. Such constraints hinder their use in the time-critical environment of medical facilities. Recently deep learning has been extensively applied in various other fields including audio \cite{latif2018phonocardiographic}, speech \cite{latif2018transfer,latif2017variational}, and vision \cite{8281753} but very few attempts have been made for motion correction in MRI. Loktyushin et al. \cite{loktyushin2015retrospective} studied the performance of convolution neural network (CNN) for retrospective motion correction in MR images. They trained the model to learn a mapping from the motion-corrupted data to motion-free images. The study indicated the potential application of deep neural networks (DNNs) to solve the motion problem in MRI, however, it lacks the detailed investigation of technique and quantitative representation of results. Similarly, Duffy et al. \cite{duffy2018retrospective} used CNN to correct motion-corrupted MR images. The work has been compared with traditional Gaussian smoothing \cite{haddad1991class} and significant improvement has been reported but comparison with the advanced state-of-the-art iterative motion correction techniques was unaccounted. Most importantly, studies on motion correction using deep learning have not exploited GANs, despite of their success in bio-medical image analysis \cite{wolterink2018generative} and modeling of natural images \cite{radford2015unsupervised,ledig2017photo} to date. However, they have been proved very powerful for MRI reconstruction \cite{mardani2017deep,yang2018dagan}. In our previous work \cite{latif2018automating}, we proposed the utilization of GAN \cite{goodfellow2014generative} for multishot MRI  motion correction. This work presented the preliminary results on motion correction by reducing the computational time greatly. However, the study has the deprivation of detailed investigation of the proposed framework for multishot MRI against the various parameters such as \textit{the number of shots} and \textit{the encoding trajectories}. We extend that work and propose an adversarial CG-SENSE reconstruction framework for the correction of the motion. A detailed analysis of the proposed framework has been presented with respect to different parameters of multishot imaging such as \textit{the amounts of motion, the number of shots,} and \textit{the encoding trajectories}.

\section{Methodology}
\label{pro}

\begin{figure}[t]
\centering
\centerline{\includegraphics[width=.5\textwidth]{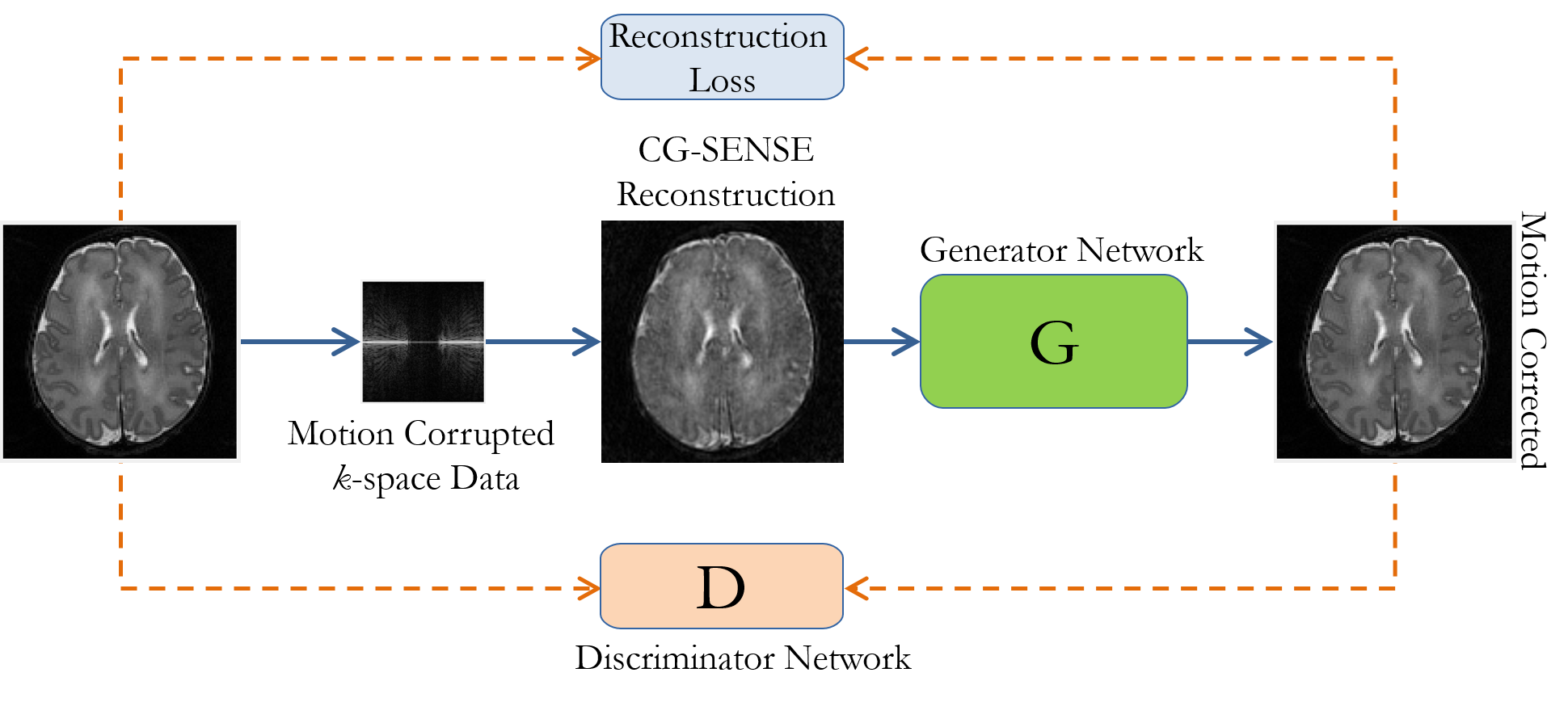}}
\caption{The proposed motion correction framework for multishot MRI, where CG-SENSE is used to reconstruct the motion-corrupted
images, and the generator network of the GAN, in conjunction with the discriminator network, is tasked with motion correction (Figure Credit: \cite{latif2018automating})}
\label{proposed}
\end{figure}

In our proposed method, reconstruction and motion correction are performed, independently. Standard CG-SENSE is employed to reconstruct \textit{k}-space data which provides the motion-corrupted image in the spatial domain. Motion-corrupted images are given to the GAN for the cleaning of motion artifacts in the second stage. Fig. \ref{proposed} shows the overall proposed architecture. 

\subsection{Motion Model for Multishot MRI}
\label{motion_model}

In Multishot MRI,  \textit{k}-space data is acquired in multiple shots (i.e., 2, 4 or 8 shots) in order to cover the whole \textit{k}-space. The MRI scanners capture Fourier coefficients along encoding trajectories which is directed by the gradient shapes of the MRI sequence. For generating motion-corrupted data, we opted the same model as followed by \cite{loktyushin2015retrospective,cordero2016sensitivity}, originally proposed in \cite{batchelor2005matrix}. 

\begin{figure}[!ht]
\centering
\centerline{\includegraphics[width=.5\textwidth]{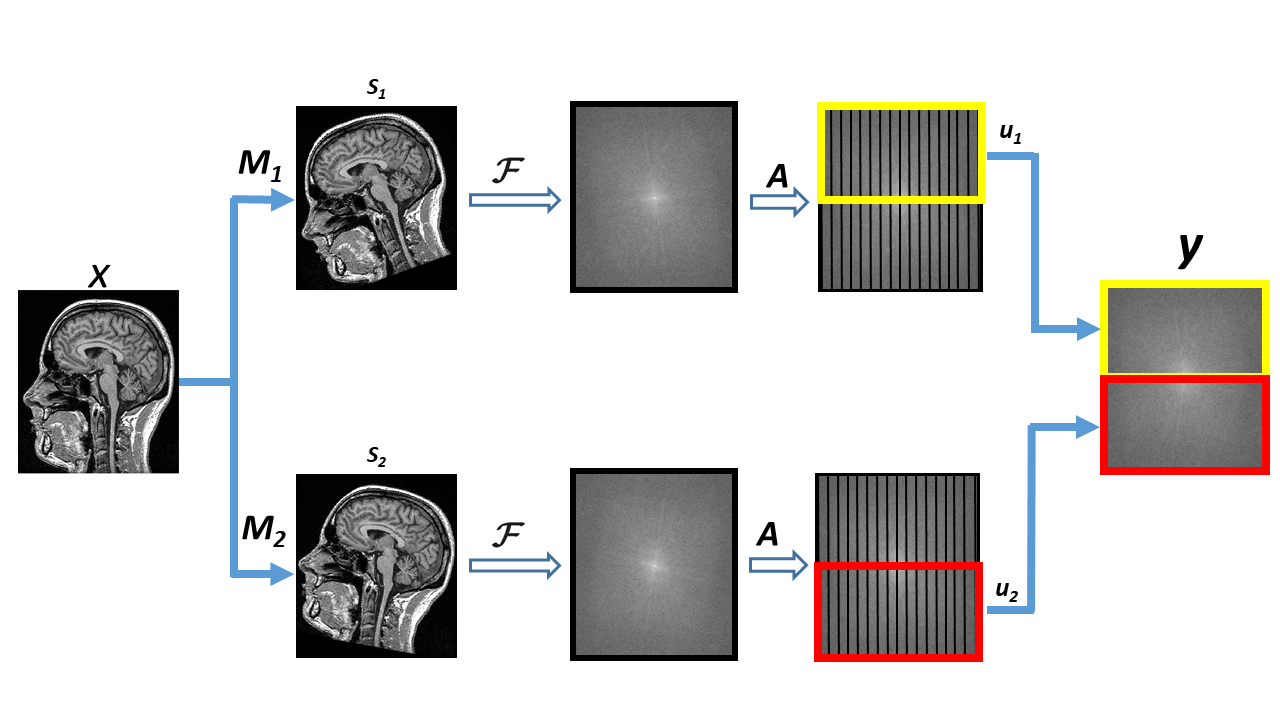}}
\caption{Forward motion corruption model (in 2D) for single coil and two shots MRI: $x$ is the motion-free image; $M_s$ is responsible for introducing the motion in particular shot; $F$ and $A$ employs DFT and sampling; and $u_s$ extracts the $k$-space segment for each shot.}
\label{data}
\end{figure}

In this model, motion $M_s$ is introduced for each $s^{th}$ shot in a motion-free image $x$. Subsequently, Fourier transform $F$ and sampling matrix $A$ is applied to achieve the \textit{k}-space representation. Finally, the segment $u_s$ of \textit{k}-space is extracted for each shot and eventually, all the segments are combined to obtain the full \textit{k}-space data. Mathematically, it can be written as:
\begin{equation}
y = \sum_{s=1}^{N} xM_sFAu_s 
\end{equation}
where, $N$ represents the number of shots, $M_s$ the translation as well rotational motion for $s^{th}$ shot, and $y$ the motion-corrupted \textit{k}-space data. Fig. \ref{data} shows the forward motion model for single coil and two shots. 
\subsection{Conjugate Gradient SENSE (CG-SENSE) Reconstruction}

In our proposed technique, we employ CG-SENSE reconstruction technique to reconstruct motion-corrupted \textit{k}-space data. It utilises conjugate gradient (CG) \cite{hestenes1952methods} algorithm to efficiently solve the SENSE equations \cite{pruessmann1999sense}, which relates the gradient encoding, sensitivities and aliased images to unaliased ones. CG-SENSE algorithm relates the object to be imaged $x_m$, the encoding matrix $E$ and the acquired \textit{k}-space data $y$ as follows:

\begin{equation}
    Ex_m = y
    \label{eq1}
\end{equation}
The acquired data $y$ has size $n_cn_k$, where $n_c$ and $n_k$ are the number of coils and the number of sampled positions in \textit{k}-space, respectively. The size of reconstructed image $x_m$ is $N^2$, while $N$ is the matrix size of the image. The spatial encoding information of gradients and coil sensitivities, is presented by the encoding matrix $E$.

To solve equation (\ref{eq1}), $E$ has to be inverted, which is a difficult task due to its large size. 
Therefore,  CG algorithm is used,  to iteratively solve equation  (\ref{eq1}) for the unaliased image, due to its fast convergence compared to other methods \cite{wright2014non}. 
To facilitate the formulation of the CG-SENSE reconstruction, another matrix $Z$ is introduced to inverse the encoding as follows:
\begin{equation}
    ZE = I_d
    \label{eq3}
\end{equation}
where, $Z$ and $I_d$ represents the reconstruction matrix and the identity matrix, respectively. Multiplying both sides of equation (\ref{eq1}) by the $F$ matrix results into an unaliased image which can be described as:

\begin{equation}
    x_m = Zy
    \label{eq4}
\end{equation}
The reconstruction matrix $Z$ can be computed by employing Moore-Penrose inversion:
\begin{equation}
    Z = (E^HE)^{-1}E^H
    \label{eq6}
\end{equation}
Now the set of equations can be solved without finding the inverse of the $E$ matrix by employing CG algorithm. To efficiently perform the CG-SENSE reconstruction process pre-conditioning is performed for better initial estimation of $x$ \cite{wright2014non}. 

\subsection{Generative Adversarial Framework}
GANs \cite{goodfellow2014generative} are latent variable generative models that learn via an adversarial process to produce realistic samples from some latent variable code. It includes a generator $G$ and a discriminator $D$ which play the following two-player min-max game:
\begin{equation}
\label{Gan}
    \underset{G}{\text{min}} \  \underset{D}{\text{max}}
    \quad
    \mathrm{E}_x[\log(D(x))] + \mathrm{E}_z[\log(1 - D(G(z)))]
\end{equation}
In a simple vanilla GAN, the generator $G$ maps the latent vectors drawn from some known prior $p_{z}$ (simple distribution e.g. Gaussian) to the sample space. The discriminator $D$ is tasked with differentiating between samples generated $G(z)$ (fake) and data samples (real). 

Here, we use conditional GAN, where instead of random samples, $G$ is fed corrupted MRI images $x_m$ and is trained to produce motion corrected image $x_c$.  The adversarial training loss $\mathcal{L}_{\text{adv}}$ for $G$ is defined as:
\begin{equation}
    \mathcal{L}_{\text{adv}}  = \log( 1 -  D(G(x_m)))
\end{equation}
To facilitate the generator, in addition to the adversarial loss, we also incorporate data mismatch term.
\begin{equation}
\mathcal{L}_{\text{data}} = \| x_c - G(x_m) \|_2
\end{equation}
Adversarial training encourages the network to produce sharp images, which is of crucial importance in MRI imaging, whereas data mismatch loss forces the network to correctly map degraded images to the original ones.
Thus the final loss for $G$, dubbed generator, is a weighted sum of $\mathcal{L}_{\text{data}}$ and $\mathcal{L}_{\text{adv}}$.
\begin{equation}
    \mathcal{L} = \mathcal{L}_{\text{data}} + \lambda\mathcal{L}_{\text{adv}}
\end{equation}
where $\lambda$ is a hyper-parameter that controls the weight of each loss term. As training progresses, $G$ and $D$ are trained iteratively.

\section{Experimental Setup}
\label{Ex}
\subsection{Dataset}
For the evaluation of the proposed method, publicly available data is utilized. T2 FLAIR images from Brain Tumor Image Segmentation (BraTS) Challenge 2015 \cite{menze:hal-00935640} dataset is used, which contains 274 scans and each scan contains 255 slices. Scans are separated into 70\% and 30\% for training and testing data, respectively. Images of BraTS dataset are considered as motion-free images and motion is introduced by employing the model described in Section \ref{data}. The same perturbation technique has been employed in \cite{cordero2018three,loktyushin2015retrospective}. As BraTS contains spatial domain images, we used a reference scan to estimate the coil sensitivity maps by using \cite{allison2013accelerated}. For our work, we produce data with varying degrees of angular motion, number of shots, and trajectories to validate the robustness of our proposed technique.

\begin{figure}
    \centering
    \includegraphics[width=.48\textwidth]{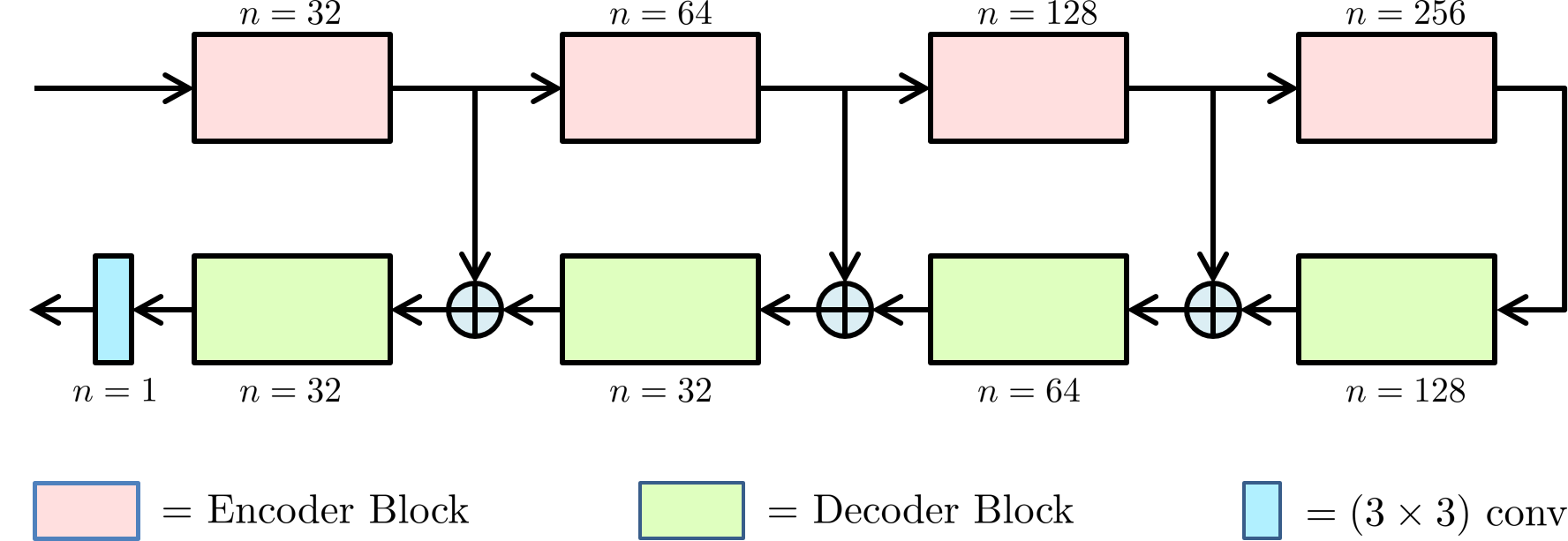}
    \caption{U-Net like Model Architecture used as Generator and Discriminator in GAN (Figure Credit: \cite{latif2018automating})}
    \label{fig:model_arch}
\end{figure}

\begin{table*}[!ht]
\caption{Performance metrics of our approach on different degree of amount of motion with 16-shots and Random trajectory}
\centering
\scalebox{1.1}{
\scriptsize
\begin{tabular}{|l|l|l|l|l|l|l|}
\hline
\textbf{Degree of motion}  & \textbf{2\degree} & \textbf{5\degree} & \textbf{8\degree} & \textbf{10\degree} & \textbf{12\degree} & \textbf{14\degree} \\ \hline
\textbf{Peak signal to noise ratio (PSNR)} & 32.31 & 31.57 & 30.89 & 28.18 & 27.85 & 27.25 \\ \hline
\textbf{Structural similarity index (SSIM)} & 0.96 & 0.96 & 0.94 & 0.92 & 0.91 & 0.90 \\ \hline
\textbf{Artifact power (AP)} & $2.47\times10^{-3}$ & $4.52\times10^{-3}$ & $6.57\times10^{-3}$ & $7.31\times10^{-3}$ & $8.08\times10^{-3}$ & $9.10\times10^{-3}$ \\ \hline
\end{tabular}
}
\label{angle}
\end{table*}

\subsection{Model Architecture}

We adopt a U-Net like architecture (shown in Figure \ref{fig:model_arch}) because of its recent success in image restoration task \cite{mao2016image} \cite{lustig2008compressed}. This involves an encoder and decoder. Due to the bottleneck in this hour-glass structure, the encoder learns to compress relevant information from the corrupted MRI scan discarding the corruption such that decoder is able to restore a clean, un-corrupted counterpart. Encoder consists of convolutions blocks, where each block consists of convolutional layers following by non-linear activation; decoder blocks are composed of transposed convolution layers. 

This hourglass structure of U-Net consists of symmetric skip connections from encoder blocks to the decoder blocks. This is necessary to recover fine details for better image restoration: encoder learns to compress image into the high-level features necessary for image restoration, but may remove fine details along with the corruptions, whereas the skip connections from encoder to decoder transfer low-level features from the encoding path to the decoding path to recover the details of the image. In addition to these skip connections, we employ residual connections inside each encoder and decoder block. These residual connections along with skip connections allow efficient gradient back-propagation, which helps in alleviating issues such as vanishing gradients and slow convergence.

The high-level model architecture is described in Fig. \ref{fig:model_arch}. Each encoder block consists of 5 convolution layers, each with $n$ feature maps except for the layer in the middle with $n/2$ feature maps. Padding is employed to keep the dimension of feature maps same inside each block. We set the strides equal to $1$ for all layers except the first one, where we choose it to be 2. This stride 2 convolution serves to down-sample feature maps using a learned kernel. Inside each encoder block, a residual connection is used between the first layer and the last layer. Decoder block has the same structure as the encoder except that we replace all the convolutional layers with transposed convolutions and use a stride of 2 at the last layer instead of the first layer. Here stride 2 transposed convolution serves to up-sample the feature maps along the U-Net architecture. The discriminator is exactly the same as the encoder part of the generator.

\subsection{Model Training}
We train our network on synthetically generated dataset using the RMSProp optimizer with the learning rate being $1\times 10^{-4}$ and a batch size of $16$, until convergence. For each update of $G$, we update $D$ twice. We pre-train the generator $G$ using Adam optimizer with same learning rate and batch size. This allows training of $G$ to converge faster, while we choose $\lambda$ to be $0.01$.

\begin{figure*}[!ht]
\centering
\centerline{\includegraphics[width=.8\textwidth]{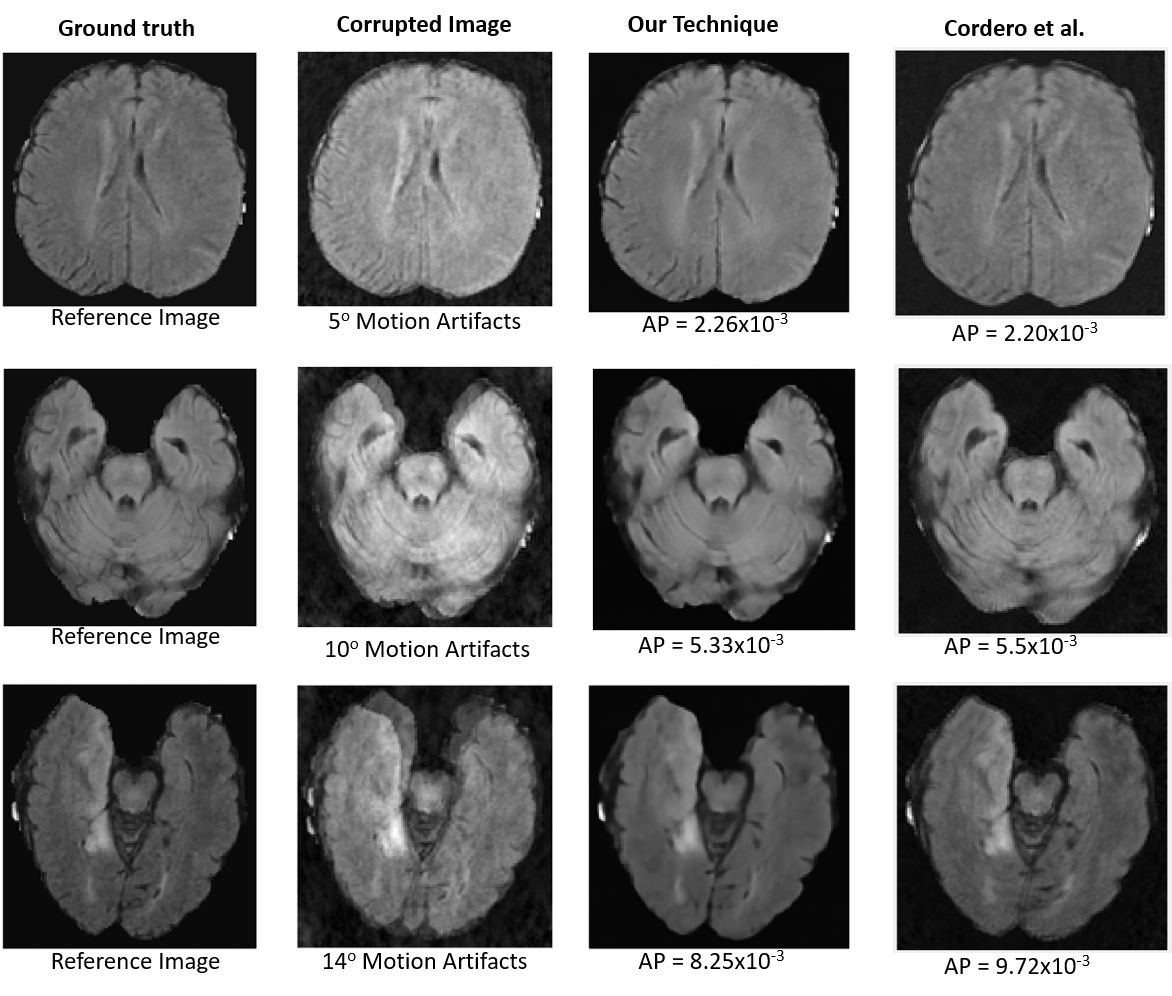}}
\caption{Resultant images of the proposed techniques and Cordero et al. \cite{cordero2018three} for $\Delta\theta = \{5\degree, 10\degree, 14\degree\}$ with 16-shot and Random Trajectory}
\label{comparison}
\end{figure*}

\section{Results and Discussion}
\label{RE}
In this section, we have performed a detailed investigation of our proposed technique for the reconstruction of motion-free images in the presence of varying amounts of motion, number of shots, and encoding trajectories. For validation, we used peak signal to noise ratio (PSNR), structural similarity Index (SSIM), and artifact power (AP) as quantification parameters. 

\subsection{Effect of the amount of motion}

\begin{table*}[!ht]
\caption{Performance metrics of our approach for varying shots at 5 degree.}
\centering
\scriptsize
\begin{tabular}{|l|l|l|l|l|l|l|l|}
\hline
\textbf{Number of Shots} & \textbf{2} & \textbf{4} & \textbf{8} & \textbf{16} & \textbf{32} & \textbf{64} & \textbf{128} \\ \hline
\textbf{Peak signal to noise ratio (PSNR)} & 31.82 & 31.92 & 31.55 & 31.57 & 31.93 & 32.02 & 32.08 \\ \hline
\textbf{Structural similarity index (SSIM)} & 0.95 & 0.96 & 0.96 & 0.96 & 0.96 & 0.96 & 0.96  \\ \hline
\textbf{Artifact power (AP)} & $4.52\times10^{-3}$ & $4.42\times10^{-3}$ & $4.32\times10^{-3}$ & $4.13\times10^{-3}$ & $3.56\times10^{-3}$ &$3.59\times10^{-3}$ & $3.46\times10^{-3}$\\ \hline
\end{tabular}
\label{shotz}
\end{table*}

To evaluate the effect of motion, different rotational motion artifacts have been introduced into motion-free images with 16-shots and random trajectory. Motion-corrupted \textit{k}-space data has been reconstructed using CG-SENSE (without motion correction) and then fed to the adversarial network, which is tasked to generate the motion-free images. Table \ref{angle} summarizes the average results obtained for varying degrees of rotational motion ($\Delta \theta = \{2\degree, 5\degree, 8\degree, 10\degree, 12\degree, 14\degree $\}) on test data. It can be noted from Table \ref{angle} that the proposed framework shows excellent performance for small amount of motion by capturing the underlying statistical properties of MR images, and recover sharp and excellent images. However, with the increase in the amount of motion, a smooth decay in the performance of model is observed, as expected, because at higher degree (i.e., 14$\degree$) MRI scans severely degraded and it becomes very difficult to recoverable the motion free image. 
Moreover, the performance of our technique is better than the previous state-of-the-art iterative technique \cite{cordero2016sensitivity} for higher amounts of motion (i.e., $\Delta\theta = 14\degree$) (see Fig. \ref{comparison}). For a small amount of motion, the approach of Cordero et al. \cite{cordero2016sensitivity} performs slightly better in terms of AP, however, the long computational time restrains its efficiency. 

\subsection{Influence of the Number of Shots}

In this experiment, we investigate the performance of the proposed framework for different number of shots. We generated motion-corrupted data for various number of shots, (i.e., S = \{2,4,8,16,32,64,128\}) with \textit{five degree} of motion and \textit{the random trajectory}. We trained our model individually for each \textit{number of shots} and evaluated the performance. The results are summarized in Table \ref{shotz}, which describes the mean values of results obtained on all the test scans. It can be seen from Table \ref{shotz} that the network is able to learn the artifact pattern and provides significantly promising results for all the number of shots. Encouragingly, our network produces sharp images with high values of PSNR and SSIM even for a higher number of shots. In contrast, state-of-the-art iterative technique \cite{cordero2016sensitivity} were only able to correct the motion for lower number of shots effectively (see Fig. \ref{shots}). For higher number of shots ($S>=32$), the convergence of such iterative techniques \cite{loktyushin2015retrospective,cordero2016sensitivity} becomes very difficult. In our case, the motion is corrected in the spatial domain after the full reconstruction of the motion-corrupted image, which enables the adversarial network to correct the motion artifacts in the image domain without encountering such convergence challenges. 

\begin{table*}[!ht]
\caption{Performance metrics of our approach for different trajectories of multishot MR imaging for eight number of shot ($S = 8$) and 5\degree of motion.}
\scriptsize
\centering
\scalebox{1.2}{
\begin{tabular}{|l|l|l|l|l|l|}
\hline
\textbf{Sampling Trajectory} & \textbf{Cartesian sequential} & \textbf{Cartesian parallel 1D} & \textbf{Cartesian parallel 2D} & \textbf{Random} \\ \hline
\textbf{Peak signal to noise ratio (PSNR)} & 30.11 & 30.14 & 30.72 & 31.55\\ \hline
\textbf{Structural similarity index (SSIM)} & 0.95 & 0.95 & 0.95 & 0.96 \\ \hline
\textbf{Artifact power (AP)} & $5.51\times10^{-3}$ & $5.425\times10^{-3}$ & $5.025\times10^{-3}$ & $4.32\times10^{-3}$ \\ \hline
\end{tabular}
}
\label{traject}
\end{table*}

\subsection{Influence of the Encoding Trajectory}
\begin{figure}[t]
\centering
\centerline{\includegraphics[width=.4\textwidth]{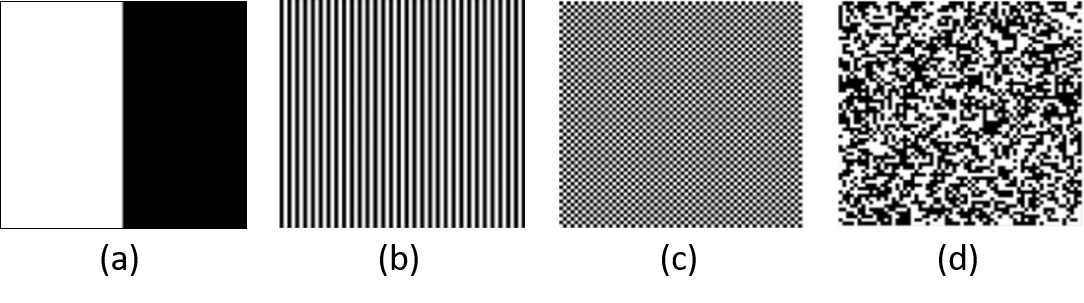}}
\caption{Encoding strategies used for experiments, depicted for S = 2 shots: (a) Cartesian sequential, (b) Cartesian parallel 1D, (c) Cartesian parallel 2D, (d) Random; samples corresponding to one of the shots is in white and those corresponding to the other is in black.}
\label{trajectories}
\end{figure}
From the vast range of trajectories, we restricted ourselves to the four trajectories (as shown in Fig. \ref{trajectories}) to validate the performance of the proposed framework. 
The motion corrupted data of each encoding trajectory is generated with eight number of shots ($S = 8$) and a relative rotation of $\Delta\theta = 5\degree$ is assumed between shots. We first performed full reconstruction of motion corrupted \textit{k}-space data for each encoding trajectory and then trained the GANs with resultant motion artifact-corrupted images, individually for each trajectory. 

\begin{figure}[!ht]
    \centering
    \subfigure[]{\includegraphics[trim=0.5cm 1.4cm 0.8cm 0.2cm,clip=true,width=.241\textwidth]{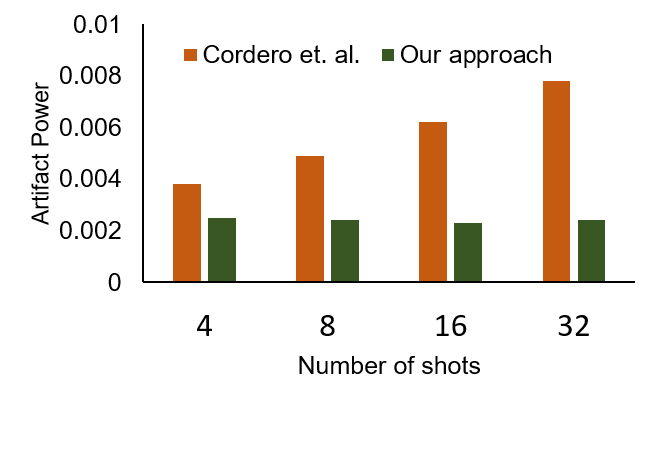}\label{shots}}
    \subfigure[]{\includegraphics[trim=0.5cm 1.4cm 0.8cm 0.2cm,clip=true,width=.241\textwidth]{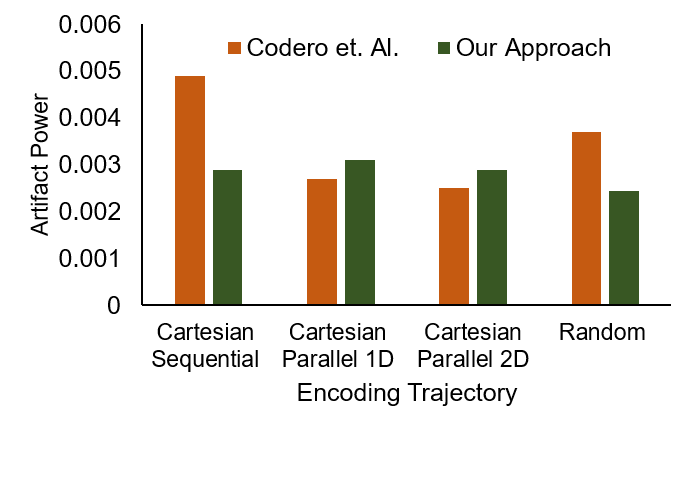}\label{tra}}
    \caption{Comparison of our framework with previous iterative technique \cite{cordero2016sensitivity} in terms of number of shorts in \ref{tra} and encoding trajectories in \ref{shots} for randomly selected fifty test images.}
    \label{fig:pk_in_coverage}
\end{figure}

Table \ref{traject} describes the mean results of our proposed framework for each encoding trajectory. The results show that our approach performs significantly well for all the encoding trajectories. However, it can be noted through a close observation that the performance of the proposed technique is slightly better for the \textit{random trajectory} since the random trajectory is least effected by the motion. The same reasoning can be applied for slightly degraded performance for Cartesian sequential trajectory as this trajectory is most affected by the motion artifacts. On the other hand, the iterative technique \cite{cordero2016sensitivity} vigorously changes its performances against different encoding trajectories (see Fig. \ref{tra}). For \textit{Cartesian sequential trajectory} this technique takes extraordinarily large number of iterations to reach the convergence, while the proposed technique has universal acceptance and it can be employed to any encoding trajectory.

\subsection{Computational time analysis}

In this section, we performed a series of experiments to evaluate the efficiency of the proposed technique in term of computational time. We compared the computational time of our technique with the previous state-of-the-art iterative technique \cite{cordero2016sensitivity}. For the sake of fair analysis, we performed the motion correction of same motion corrupted \textit{k}-space data on the same hardware by employing both techniques. Intel\textsuperscript{\textregistered} Core\textsuperscript{TM} i3-2120 CPU with 3.5GHz speed, 16GB of memory and NVIDIA\textsuperscript{\textregistered} Quadro M5000 Graphic Processing Unit (GPU) with 8GB GDDR5 memory, has been used for our experiments. The proposed technique involves two steps, i.e., CG-SENSE reconstruction and motion correction. Therefore, to calculate the total computational time, we added the reconstruction and motion correction time. Table \ref{tym_shots} provides a summary of the results comparing the computational time analysis of our technique with that proposed by Cordero et al. \cite{cordero2016sensitivity} for varying number of shots for fifty randomly selected test images. It can be seen that our technique is several times faster than the previous iterative approach \cite{cordero2016sensitivity}. The previous technique is an iterative method that first iteratively estimates the motion and then corrects for that motion, which needs extra computational time. With the increase of number of shots, it becomes difficult to estimate the motion between two consecutive shots, subsequently, it further increases the time required to correct the motion for higher numbers of shots. Moreover, changing the encoding trajectory also significantly effects the computational performance of conventional iterative technique \cite{cordero2016sensitivity}. Alternatively, in our proposed technique, motion correction is independent of the reconstruction process and it is performed after full reconstruction of \textit{k}-space data. Therefore, the motion correction for all the number of shots takes the same computational time. However, the CG-SENSE reconstruction takes more time for higher number of shots, which slightly increases the overall motion corrected reconstruction time (see Table \ref{tym_shots}). In Table \ref{tym}, we summarize the computational time of our technique and iterative technique \cite{cordero2016sensitivity}, against different amounts of motion. The time required to correct for motion in our technique is not dependent upon the amount of motion, therefore, it remains the same for all amounts of motion. Alternatively, the conventional technique  takes longer time to estimate the higher amount of motion, thus it takes more time to correct such motion.

\begin{table}[!ht]
\caption{Comparing computational time in seconds of our approach with the current state-of-art technique \cite{cordero2016sensitivity} for various amounts of motion.}
\centering
\scriptsize
\begin{tabular}{|l|l|l|l|l|l|}
\hline
\textbf{Degree of motion} & \textbf{5\degree} & \textbf{8\degree} & \textbf{10\degree} & \textbf{12\degree} & \textbf{14\degree}\\ \hline
\textbf{Cordero et al.} & 25.23 & 39.27 & 57.93 & 134.27 & 152.03\\ \hline
\textbf{Our approach} & 0.28 & 0.28& 0.28 & 0.28 & 0.28 \\ \hline
\end{tabular}
\label{tym}
\end{table}

\begin{table}[!ht]
\scriptsize
\caption{Comparing computational time in seconds of our approach with the current state-of-art technique \cite{cordero2016sensitivity} against different number of shots.}
\centering
\begin{tabular}{|l|l|l|l|l|l|l|}
\hline
\textbf{Number of shots} & \textbf{4} & \textbf{8} & \textbf{16} & \textbf{32} & \textbf{64} & \textbf{128}\\ \hline
\textbf{Cordero et al.} & 9.00 & 12.80 & 14.59 & 27.83 & 50.07 & 89.19 \\ \hline
\textbf{Our approach} & 0.23 & 0.28 & 0.34 & 0.48 & 0.79 & 1.40 \\ \hline
\end{tabular}
\label{tym_shots}
\end{table}

\section{Conclusions}
\label{co}
We introduced a flexible yet robust retrospective motion correction technique that employs generative adversarial networks (GANs) to correct motion artifacts in multishot Magnetic Resonance Imaging (MRI). This work is an extension of our previous preliminary work, where we empirically showed the suitability of GAN for motion correction in multishot MRI. The proposed technique first performs the full reconstruction of motion-corrupted \textit{k}-space data and then the resultant artifact-affected image is fed into the deep generative networks that learns the mapping from motion artifact-affected images to the artifacts free images. Our GAN based framework removes the motion artifacts without any prior estimation of motion during the data acquisition or reconstruction process in contrast to the previous iterative methods. Such parameter-free technique can be employed to any encoding scheme without introducing modifications in the acquisition sequence. To validate our method, we carried out a comprehensive experimentation by varying different parameters, such as different levels of motion, the number of shots, and encoding schemes, of multishot MRI. Based on the results, we demonstrated that the performance of the proposed technique is more robust against these parameters and it also reduced the computational time significantly in contrast to the state-of-the-art techniques. Future plans include the extension of framework to perform end-to-end learning using generative network from motion corrupted \textit{k}-space data to artifacts free image.






\end{document}